\documentclass{article}

\usepackage[preprint, nonatbib]{tackling_climate_workshop_style}

\usepackage[utf8]{inputenc} 
\usepackage[T1]{fontenc}    
\usepackage{hyperref}       
\usepackage{url}            
\usepackage{booktabs}       
\usepackage{amsfonts}       
\usepackage{nicefrac}       
\usepackage{microtype}      

\usepackage[nocompress]{cite}
\usepackage{amsmath}
\usepackage{algorithmic}
\usepackage{algorithm}
\usepackage{array}
\usepackage[caption=false,font=normalsize,labelfont=sf,textfont=sf]{subfig}
\usepackage{textcomp}
\usepackage{stfloats}
\usepackage{verbatim}
\usepackage{graphicx}
\usepackage{cite}
\usepackage{mathrsfs}%
\usepackage{xcolor}%
\usepackage{manyfoot}%
\usepackage{listings}%
\usepackage{times}
\usepackage{epsfig}
\usepackage{longtable}
\usepackage{amssymb}
\usepackage{sidecap}
\usepackage{bbm}
\usepackage{multirow}
\usepackage{multicol}
\usepackage{wrapfig}
\usepackage{xspace}

\title{Efficient Localized Adaptation of Neural Weather Forecasting: A Case Study in the MENA Region}

\author{\textbf{Muhammad Akhtar Munir}$^{1}$~, \textbf{Fahad Shahbaz Khan}$^{1, 3}$, ~\textbf{Salman Khan}$^{1, 2}$  \\
\small{$^{1}$Mohamed bin Zayed University of AI,
}
\small{$^{2}$Australian National University, $^{3}$Linköping University
}\\
\texttt{akhtar.munir@mbzuai.ac.ae}
}

\begin{document}

\maketitle
\begin{abstract}
Accurate weather and climate modeling is critical for both scientific advancement and safeguarding communities against environmental risks.
Traditional approaches rely heavily on Numerical Weather Prediction (NWP) models, which simulate energy and matter flow across Earth's systems. 
However, heavy computational requirements and low efficiency restrict the suitability of NWP, leading to a pressing need for enhanced modeling techniques. 
Neural network-based models have emerged as promising alternatives, leveraging data-driven approaches to forecast atmospheric variables. 
In this work, we focus on limited-area modeling and train our model specifically for localized region-level downstream tasks. 
As a case study, we consider the MENA region due to its unique climatic challenges, where accurate localized weather forecasting is crucial for managing water resources, agriculture and mitigating the impacts of extreme weather events. 
This targeted approach allows us to tailor the model's capabilities to the unique conditions of the region of interest.
Our study aims to validate the effectiveness of integrating parameter-efficient fine-tuning (PEFT) methodologies, specifically Low-Rank Adaptation (LoRA) and its variants, to enhance forecast accuracy, as well as training speed, computational resource utilization, and memory efficiency in weather and climate modeling for specific regions. Our codebase and  pre-trained models can be accessed at \url{https://github.com/akhtarvision/weather-regional}.
\end{abstract}

\section{Introduction}
The accurate modeling and prediction of weather and climate patterns hold prominent significance for both scientific research and societal well-being. 
Primarily, weather and climate modeling can be categorized into two types: numerical (NWP) methods and neural network-based models. The former, usually related to General Circulation Models (GCMs) \cite{lynch2008origins, bauer2015quiet}, are designed to simulate the flow of energy and matter across the land, atmosphere, and ocean. 
However, the ability to perform detailed simulations to forecast weather and atmospheric variables in NWP models is restricted by computational resources and the time required for execution.
As a consequence, there is a pressing need to address computational challenges and improve the precision of weather and climate modeling.

Neural network-based models \cite{bi2023accurate, dueben2018challenges, grover2015deep, lam2023learning, nguyen2023climax, pathak2022fourcastnet, kurth2023fourcastnet, ben2023rise, scher2019weather, schultz2021can, weber2020deep} are data-centric and can absorb large-scale data to enhance performance. 
These models undergo comprehensive training on globally accessible datasets, thereby providing a viable alternative for predicting weather and climate patterns. 
Despite the absence of explicit physics assumptions in these models, the datasets employed for predictive task training inherently incorporate implicit physics assumptions. 
Within data-driven approaches, transformer-based \cite{nguyen2023climax, bi2023accurate} and graph-based \cite{lam2023learning} models have been explored. 
One notable example of a transformer-based model is ClimaX \cite{nguyen2023climax}, which focuses on developing a foundational model for climate and weather applications. 
ClimaX introduces a pre-training and fine-tuning paradigm, where during pre-training, it trains on data curated from physics-based models, enhancing its predictive capabilities and capturing relationships among atmospheric variables. This is followed by finetuning for various climate and weather tasks on a comprehensive climate reanalysis dataset.
Other prominent methods include Pangu \cite{bi2023accurate} and GraphCast \cite{lam2023learning}, with the latter being a graph neural network based approach for weather prediction. GraphCast operates on a high-resolution latitude-longitude grid, employing an architecture consisting of an encoder, a processor, and a decoder. 
However, these models focus on global forecasts and do not account for region-specific dynamics.

\begin{figure*}[t]
    \centering
    \includegraphics[width=1.0\linewidth]{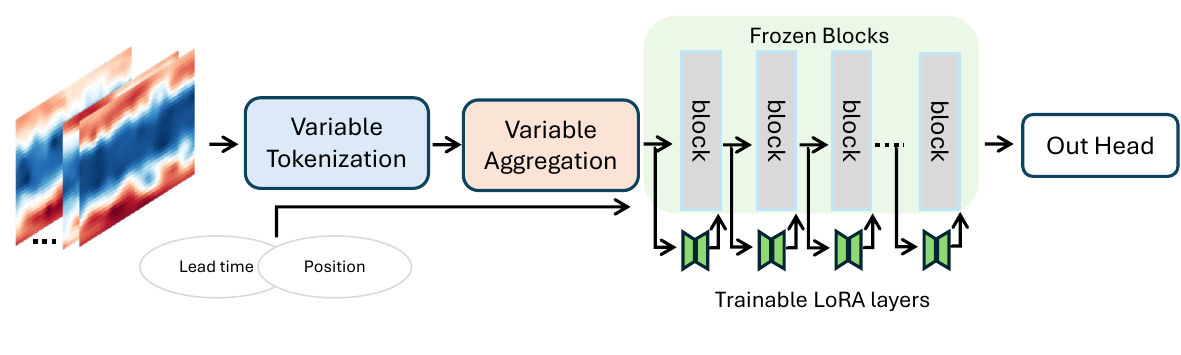}
    \caption{\small {Main architecture:} Integration of LoRA involves trainable layers while transformer blocks are frozen. 
    The architecture modifies the main ViT by dealing with each channel separately for tokenization. 
    }
    \label{fig:main}
\end{figure*}

Our study focuses on a transformer-based weather and climate forecasting model trained for 
localized regional forecasting. The model capitalizes on comprehensive fine-tuning of the pre-trained weights of the global foundational model. It is crucial to investigate how these models can adapt to a paradigm of parameter-efficient fine-tuning (PEFT) \cite{hu2021lora, lester2021power, jia2022visual, chavan2023one}, providing efficient training 
and maintaining optimal performance. This paper implements various PEFT methods, primarily incorporating Low Rank Adaptation (LoRA) and its variants, over the ClimaX model. 
LoRA enhances model efficiency by introducing trainable components to each layer, thereby reducing the number of trainable parameters in large-scale models (Fig. \ref{fig:main}). This approach addresses challenges related to speed, computational resources, and memory efficiency during the training of large models.

In the pursuit of adapting large-scale models with a parameter-efficient tuning methodology for downstream forecasting tasks, our proposal involves the exploration and investigation of LoRA and its variants across the Middle East and North Africa (MENA).
The MENA region has experienced significant climate impacts, with summer temperatures projected to rise at more than twice the global average. This increase in heat, along with extended heat waves, could affect the region's habitat and can affect human health \cite{lelieveld2016strongly}. 
We aim to corroborate the efficacy of the proposed strategies in enhancing performance metrics, speed, and memory efficiency. Additionally, we integrate the flash attention mechanism into our approach, desirably accelerating the attention calculation process, and thereby facilitating resource-efficient model training for localized regions.


\section{Method}\label{sec3}

\subsection{Preliminaries}\label{subsecm1}

\noindent\textbf{Notations:} To input the neural network model, it takes the input $\mathcal{I}$ of shape $D \times H \times W$, where $D$ is the number of variables spanning the atmospheric or climate ones. 
Let $\mathcal{F}$ be a neural network operator for gridded prediction tasks which takes input, $\mathcal{F} (\mathcal{I})$ and output $\mathcal{O}$ of shape $\hat{D} \times \hat{H} \times \hat{W}$. Spatial resolution $H \times W$ determines the density of the grid, and here we operate with two levels of resolutions 5.625$^{\circ}$ (32 $\times$ 64 grid points) and 1.40625$^{\circ}$ (128 $\times$ 256 grid points).

\noindent\textbf{Architecture:} 
The Vision Transformer (ViT) architecture involves the partitioning of input data into fixed patches, followed by a linear transformation to generate patch embeddings, commonly referred to as tokens \cite{dosovitskiy2020image}. 
The ClimaX\cite{nguyen2023climax} model we use consists of two main features, which will be briefly described in this paragraph. \textbf{(i) Variable Tokenization:} The proposed framework aims to mitigate a limitation observed in ViT architecture, which inherently processes a specified number of channels in the input. 
This approach independently addresses each input atmospheric and climate variable.
\textbf{(ii) Variable Aggregation:}  
The variable tokenization approach presents challenges, notably an increase in sequence length corresponding to the rise in atmospheric and climate variables. Consequently, the memory complexity escalates quadratically when containing attention mechanisms. 
To address these challenges, a cross-attention operation is implemented for each spatial position within the map. 
This strategy effectively mitigates the computational complexity.
For more details, we refer readers to \cite{nguyen2023climax}.

\subsection{Fine tuning using PEFT}\label{subsecm2}

Several methods are considered under the PEFT paradigm. We extensively study LoRA and its variants along with GLoRA, to cater to several PEFT mechanisms in one place. 

\noindent\textbf{LoRA:} It presents several advantages, including the capability to integrate multiple LoRA modules for specific tasks without modifying the underlying base foundation model. By freezing the weights of the base model and optimizing low-rank learnable matrices, significant reductions in storage requirements can be accomplished. Additionally, there is no added inference latency compared to using a fully fine-tuned model. Because of these advantages, we also explore a variant of LoRA that incorporates residuals to retain information from previous blocks. However, empirical observations indicate that the simple LoRA method continues to outperform these variants.
Neural network layers typically possess full rank, while LoRA attempts to project weights into a smaller subspace. Let $\mathcal{W}_p$ represent the pretrained weight matrix, and its modification $\Delta\mathcal{W}$ is replaced by low-rank decomposed matrices, denoted as $BA$, resulting in the equation:
\begin{equation}
    \mathcal{W}_p + \Delta\mathcal{W} = \mathcal{W}_p + BA 
    \label{Eq:lora_formula}
\end{equation}
Here, $B \in \mathbb{R}^{d \times r}$ and $A \in \mathbb{R}^{r \times k}$ decompose the matrix $\mathcal{W}_p \in \mathbb{R}^{d \times k}$, where $r$ is the rank determined as $r \le \min(d, k)$. To explain further, it is notable that gradients are not updated for $\mathcal{W}_p$. For more details, interested readers are encouraged to consult \cite{hu2021lora}.

\noindent\textbf{GLoRA:} 
This methodology presents a unified framework that integrates fine-tuning approaches within a singular formulation. The architecture features a supernet, which is efficiently optimized through evolutionary search techniques. These conventional methods often rely on resource-intensive hyperparameter searches, dependent upon data availability. Employing an implicit search mechanism prevents the requirement for manual hyperparameter tuning, relaying the simultaneous increase in training time. We refer to App. \ref{appglora} for more details on GLoRA.

\noindent\textbf{Flash-attention:}
In advancing large models like language models, the flash attention algorithm, introduced by \cite{dao2022flashattention}, optimizes attention computation by reducing memory usage. Its successor, flash attention-2 \cite{dao2023flashattention2}, further improves efficiency by minimizing non-matrix multiplication operations and parallelizing over sequence length.



\section{Experiments and Results}
\label{sec4}
The ERA5 reanalysis, developed by the European Center for Medium-Range Weather Forecasting (ECMWF), is utilized as a fundamental data source for training and evaluating weather forecasting systems. 
ERA5 integrates state-of-the-art Integrated Forecasting System \cite{wedi2015modelling} model outputs with observational data to generate comprehensive records of atmospheric, and land surface conditions. 
More details regarding data and implementation can be found in the App. \ref{appdata} and App. \ref{appimplement} respectively.

\begin{figure*}[b]
    \centering
    \includegraphics[width=1.0\linewidth]{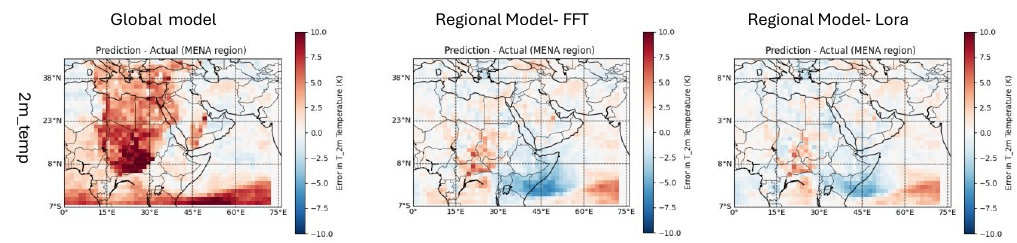}
    \caption{\small {Qualitative: Error/Bias in Predictions and Actual measurements for temperature\_2m (K)}. Dated, 11$^{th}$ April 2017, lead time 3 days.
    }
    \label{fig:climaxvis}
\end{figure*}

\begin{table*}[t]
\fontsize{7.5}{10}\selectfont
\centering
\renewcommand{\arraystretch}{1.1}
\tabcolsep=1.5pt\relax
\begin{tabular}{c|c|c|c|c|c|c|c|c}
\hline
\textbf{Metric}                & \textbf{Model} & \textbf{geop@500} & \textbf{2m\_temp} & \textbf{r\_hum@850} & \textbf{s\_hum@850} & \textbf{temp@850} & \textbf{10m\_u\_wind} & \textbf{10m\_v\_wind} \\ \hline
\multirow{2}{*}{\textbf{ACC} ($\uparrow$)}  & Global         & 0.292            & 0.230            & 0.255              & 0.282              & 0.246            & 0.287                & 0.238                \\  
                               & Regional       & 0.585           & 0.804            & 0.502              & 0.623              & 0.620            & 0.570                & 0.517                \\ \hline
\multirow{2}{*}{\textbf{RMSE} ($\downarrow$)} & Global         & 674.295          & 3.349            & 23.308             & 0.003              & 3.561            & 3.733                & 4.162                 \\  
                               & Regional       & 411.125          & 1.518            & 18.945             & 0.002              & 2.366            & 2.931                & 3.219                \\ \hline
\end{tabular}
\caption{\small \textbf{\textit{Global vs Regional}}:
Forecasting on MENA region for 72 hrs prediction. Resolution is $1.40652^\circ$. The global model performs worse whereas regional model, specific to the needs of the local region performs better.
}
\label{tab:reg_vsglobal}
\end{table*}

\begin{table*}[t]
\fontsize{7.5}{10}\selectfont
\centering
\renewcommand{\arraystretch}{1.1}
\tabcolsep=1.5pt\relax
\begin{tabular}{l|ccc|ccc}
\hline
\textbf{\textbf{Metric}}     & \multicolumn{3}{c|}{\textbf{ACC} ($\uparrow$)}                                        & \multicolumn{3}{c}{\textbf{RMSE} ($\downarrow$)}                                       \\ \hline
\textbf{Model}                &  \multicolumn{1}{c|}{geop@500} & \multicolumn{1}{c|}{2m\_temp} & temp@850 & \multicolumn{1}{c|}{geop@500} & \multicolumn{1}{c|}{2m\_temp} & temp@850 \\ \hline
\textbf{Global}                & \multicolumn{1}{c|}{0.276}     & \multicolumn{1}{c|}{0.579}     & 0.316     & \multicolumn{1}{c|}{615.428}    & \multicolumn{1}{c|}{2.111}     & 3.343     \\ \hline
\textbf{Regional$_{fft}$}      & \multicolumn{1}{c|}{0.554}     & \multicolumn{1}{c|}{0.816}     & 0.597     & \multicolumn{1}{c|}{453.256}    & \multicolumn{1}{c|}{1.448}     & 2.536     \\ \hline
\textbf{Regional$_{Lora}$}    & \multicolumn{1}{c|}{0.582}     & \multicolumn{1}{c|}{0.823}     & 0.614     & \multicolumn{1}{c|}{438.213}    & \multicolumn{1}{c|}{1.414}     & 2.476     \\ \hline
\textbf{Regional$_{resLora}$}  & \multicolumn{1}{c|}{0.580}     & \multicolumn{1}{c|}{0.823}     & 0.613     & \multicolumn{1}{c|}{438.669}    & \multicolumn{1}{c|}{1.411}     & 2.476     \\ \hline
\textbf{Regional$_{GLora}$}    & \multicolumn{1}{c|}{0.557}     & \multicolumn{1}{c|}{0.806}     & 0.591     & \multicolumn{1}{c|}{445.437}    & \multicolumn{1}{c|}{1.482}     & 2.535     \\ \hline
\end{tabular}
\caption{\small \textbf{\textit{Lora, variants and fft}}:
Forecasting on MENA region for 72 hrs prediction. Resolution is $5.652^\circ$. Lora gives a competitive performance with a full fine-tune (fft) version and also results are reported with our proposed variant of Lora (resLora) and generalized Lora (GLora). 
}
\label{tab:reg_loravar}
\end{table*}

\subsection{Experiments with Global and Regional Modeling}\label{subsece2}
To evaluate our forecasting models, we employ the ERA5 dataset, which provides global atmospheric reanalysis data at different resolutions. 
We compare ClimaX with the settings of global forecasting as well as in regional forecasting. We assess performance at both 5.625$^{\circ}$ and 1.40625$^{\circ}$ resolutions to understand the impact of spatial granularity on forecasting accuracy.
The performance of each model is evaluated using latitude-weighted root mean squared error (RMSE) and latitude-weighted anomaly correlation coefficient (ACC), standard metrics in weather prediction literature, reflecting forecast accuracy and consistency with observed data.
In addition to global forecasting, we extensively extend our analysis to regional forecasting focusing on Middle East and North Africa (MENA). 
By selecting a regional dataset (ERA5-MENA) with the same set of variables, we assess ClimaX's ability to forecast weather conditions specifically within this region. 
We compare regional ClimaX with PEFT paradigms and a global version of ClimaX. 
Furthermore, we explore the impact of training ClimaX on data at different resolutions and evaluate its performance in regional forecasting tasks.

\noindent\textbf{Regional vs Global: } 
A comparative analysis between global and regional models unveils their respective behaviors and notably, the regional model consistently demonstrates superior performance when tasked with regional prediction objectives (Table~\ref{tab:reg_vsglobal}). The regional model is specialized to predict more accurate forecasts due to the localized features learned during training.
\noindent\textbf{Comparison with regional model variants: } In Table \ref{tab:reg_loravar}, our findings demonstrate that ClimaX with LoRA shows superior performance as compared to the global model and competitive performance compared to the full fine-tune method in predicting key atmospheric variables indicating its efficacy in weather forecasting applications. 
The number of trainable parameters is reduced from 108M (fft) to 16.2M (LoRA).
Our study also presents findings obtained using GLora and a proposed modified version of Lora, which integrates residuals from similar blocks while aggregating information into subsequent transformer blocks. 
However, our observations indicate that while it does not exhibit superior performance, it does deliver reasonable results. 
We show qualitative results in Fig. \ref{fig:climaxvis} for one of the atmospheric variables. Bias is referred to as the difference between prediction and ground truth.
We show further experiments in App.\ref{appmoreres}, which include predictions over ranges and ablations, primarily focusing on the impact of rank r, LoRA’s significance on attention and FC layers, memory and time comparison, and more qualitative results.
\section{Conclusion}\label{sec5}
Neural network based models offer a promising alternative as these data-driven approaches, such as transformer-based models, leverage comprehensive training on large-scale datasets to forecast atmospheric variables with reasonable accuracy. 
Our study focuses on enhancing the transformer-based forecasting model, particularly in the Middle East and North Africa (MENA) region, through parameter-efficient fine-tuning methods such as Low-Rank Adaptation (LoRA). 
By investigating the efficacy of LoRA and its variants, alongside the integration of flash attention mechanisms, we 
 improve model prediction accuracy and efficiency in terms of speed and memory. 
 This effort represents an initial step towards efficiently adapting foundational weather models for localized regional dynamics. 

\medskip

\newpage
\bibliographystyle{IEEEtran}
\bibliography{natbib}

\newpage

\appendix
\section{More details on GLoRA}
\label{appglora}
GLoRA overall formulates in Eq.\ref{Eq:Glora_formula}

\begin{equation}
    \mathcal{G} = (\mathcal{W}_0 + \mathcal{W}_0 U + V) x + X \mathcal{W}_0 + Y b_0 + Z +b_0
    \label{Eq:Glora_formula}
\end{equation}

The tensors denoted as U, V, X, Y, and Z serve as trainable support structures for downstream tasks. Notably, during the entire fine-tuning process, the $W_0$ and $b_0$ remain fixed. Among these tensors, U is responsible for scaling the weights, while V plays a pivotal role in both scaling the input and shifting the weights. X functions as a layer-wise prompt, akin to the prompt tuning paradigm. Y and Z are used in scaling and shifting the bias, respectively, contributing to the overall functionality of the GLoRA model. For more technical details, we refer the reader to \cite{chavan2023one}.

\section{Dataset}
\label{appdata}
This dataset, available on a global 0.25$^{\circ}$ × 0.25$^{\circ}$ latitude-longitude grid, spans around 40 years, with hourly measurements encompassing 37 altitude levels and the Earth's surface. Comprising 721 × 1440 grid points, the dataset presents altitude levels in terms of pressure levels. ERA5 represents the fifth generation of ECMWF reanalysis, and offers improved spatial and temporal resolution. 
Reanalysis employs data assimilation techniques to combine model forecasts with observations, yielding a consistent and complete dataset that spans multiple decades. The incorporation of ensemble-based uncertainty estimates enhances the utility of ERA5 for climate-related applications. Furthermore, pre-calculated monthly-mean averages facilitate analysis and interpretation. This comprehensive dataset highlights the critical role of reanalysis in advancing weather modeling and climate analysis. 

\section{Implementation Details}
\label{appimplement}
In our experiments, we operate with two levels of resolutions 5.625$^{\circ}$ (32 $\times$ 64 grid points) and 1.40625$^{\circ}$ (128 $\times$ 256 grid points). 
The dataset is partitioned into training data spanning from 1979 to 2015, validation data for the year 2016, and test data covering 2017 and 2018. 
All the default variables have been utilized as inputs in our experiments. For a 5.625$^{\circ}$ resolution, a patch size of 2, and for 1.40625$^{\circ}$ resolution, a patch size of 4 is used. The learning rate adopted for our experiments is 1.0x10$^{-5}$. The embedding dimension is set to 1024. The training process is conducted using four V100 GPUs, leveraging fp16 floating point precision. 
We present the results for a wide range of atmospheric variables, including \textit{geopotential\_500, temperature\_850, 2m\_temperature, 10m\_u\_component\_of\_wind, 10m\_v\_component\_of\_wind, relative\_humidity\_850, and specific\_humidity\_850}.
We present results using the anomaly correlation coefficient (ACC) and root mean square error (RMSE). For ACC, higher values indicate better performance, while for RMSE, lower values are preferable. These metrics together provide a comprehensive assessment of model performance.
The regional setting is the MENA region which is a subset of global grid points.

With the utilization of a comprehensive set of input variables derived from atmospheric and surface data, we focus on forecasting future weather conditions given the current atmospheric state.
Specifically with a total of 48 input features, majorly includes geopotential, temperature, wind components, relative humidity, and specific humidity at various pressure levels are crucial for weather prediction tasks due to their direct influence on atmospheric dynamics.
The forecasting tasks involve predicting seven target variables: geopotential at 500hPa, 2-meter temperature, and eastward \& northward components of the 10m wind, and temperature, relative humidity, \& specific humidity at 850hPa, 
Lead times ranging from 12 hours to 72 hours are considered, encompassing multiple range forecasting scenarios.
For training the deep learning models, as described in ClimaX, we utilize a latitude-weighted mean squared error (MSE) loss function and implement early stopping based on validation loss to prevent overfitting. 
For more details, we refer readers to \cite{nguyen2023climax}.

\section{More Results}
\label{appmoreres}
\noindent\textbf{Predictions over ranges: } 
For more accurate predictions, it was observed that a specialized model with a specific lead time stands out. With this approach, we train multiple models for specific lead time ranges from 12 hours to 3 days. In Table \ref{tab:reg_ranges}, we report results on different ranges of predictions that show the behavior of atmospheric variables over time. We can observe that as the time varies from short to medium, it becomes very crucial to predict the accurate forecast.

\subsection{Ablations}\label{subsece3}
In this section, we delve into two different mechanisms. Firstly, we explore the influence of the rank number on model behavior to understand how it changes with varying rank numbers. Secondly, we investigate the integration of LoRA with components other than the attention module, aiming to assess the impact of such integration on model performance.

\noindent\textbf{Impact of rank $r$:} In investigating the impact of rank $r$ in the Lora module, we have observed that within the range of Lora ranks, optimal results are achievable by integrating rank 16. 
Specifically, in the context of low-resolution settings, our examination of rank $r$ behavior for regional forecasting over a 72-hour prediction range is shown in Table \ref{tab:reg_ranks}.

\begin{table*}[t]
\fontsize{7.5}{10}\selectfont
\centering
\renewcommand{\arraystretch}{1.1}
\tabcolsep=1.5pt\relax
\begin{tabular}{c|c|c|c|c|c|c|c|c}
\hline
\textbf{Metric}                & \textbf{Lead time} & \textbf{geop@500} & \textbf{2m\_temp} & \textbf{r\_hum@850} & \textbf{s\_hum@850} & \textbf{temp@850} & \textbf{10m\_u\_wind} & \textbf{10m\_v\_wind} \\ \hline\hline
\multirow{6}{*}{\textbf{ACC}}  & \textbf{12}        & 0.985            & 0.917            & 0.870              & 0.899              & 0.954            & 0.950                & 0.952                \\  
                               & \textbf{24}        & 0.951            & 0.914            & 0.797              & 0.846              & 0.927            & 0.907                & 0.903                \\  
                               & \textbf{36}        & 0.876            & 0.880            & 0.705              & 0.782              & 0.868            & 0.829                & 0.815                \\  
                               & \textbf{48}        & 0.779            & 0.868            & 0.631              & 0.727              & 0.792            & 0.742                & 0.712                \\  
                               & \textbf{60}        & 0.674            & 0.814            & 0.554              & 0.668              & 0.696            & 0.650                & 0.602                \\  
                               & \textbf{72}        & 0.585           & 0.804            & 0.502              & 0.623              & 0.620            & 0.570                & 0.517                \\ \hline
\multirow{6}{*}{\textbf{RMSE}} 
& \textbf{12}        & 87.689           & 0.961            & 10.721             & 0.001              & 0.877            & 1.077                & 1.125                \\  
                               & \textbf{24}        & 152.173          & 0.973            & 13.151             & 0.001              & 1.090            & 1.465                & 1.579                \\  
                               & \textbf{36}        & 237.892          & 1.223            & 15.450             & 0.001              & 1.412            & 1.957                & 2.138                \\  
                               & \textbf{48}        & 311.773          & 1.258            & 16.939             & 0.002              & 1.800            & 2.357                & 2.604                \\  
                               & \textbf{60}        & 370.279          & 1.489            & 18.206             & 0.002              & 2.142            & 2.692                & 2.982                \\  
                               & \textbf{72}        & 411.125          & 1.518            & 18.945             & 0.002              & 2.366            & 2.931                & 3.219                \\ \hline
\end{tabular}
\caption{\small \textbf{\textit{Regional models over different prediction ranges to show trends regarding lead time with respective models.}}:
Forecasting on MENA region at resolution $1.40652^\circ$. Results reported in anomaly correlation coefficient (ACC) and root mean square error (RMSE).}
\label{tab:reg_ranges}
\end{table*}

\begin{table}[t]
\fontsize{7.5}{10}\selectfont
\centering
\renewcommand{\arraystretch}{1.1}
\tabcolsep=1.5pt\relax
\begin{tabular}{c|c|c|c|c}
\hline
\textbf{Metric}                & \textbf{Rank (r)} & \textbf{geop@500} & \textbf{2m\_temp} & \textbf{temp@850} \\ \hline
\multirow{6}{*}{\textbf{ACC} ($\uparrow$)}  & 2                 & 0.575              & 0.824              & 0.612              \\  
                               & 4                 & 0.576              & 0.824              & 0.612              \\  
                               & 8                 & 0.575              & 0.824              & 0.612              \\  
                               & 16                & 0.582              & 0.823              & 0.614              \\  
                               & 32                & 0.579              & 0.823              & 0.612              \\  
                               \hline
\multirow{6}{*}{\textbf{RMSE} ($\downarrow$)} & 2                 & 439.717             & 1.400              & 2.482              \\  
                               & 4                 & 439.969             & 1.401              & 2.483              \\  
                               & 8                 & 440.129             & 1.395              & 2.479              \\  
                               & 16                & 438.213             & 1.414              & 2.476              \\  
                               & 32                & 439.885             & 1.417              & 2.482              \\  
                               \hline
\end{tabular}
\caption{\small \textbf{\textit{Impact of rank (r) in LoRA module}}:
Forecasting on MENA region with the regional model for 72 hrs prediction range. Resolution is $5.625^\circ$. 
This table shows the trends that how rank value impacts the overall performance. We report three atmospheric variables here.}
\label{tab:reg_ranks}
\end{table}

\noindent\textbf{Lora's significance on attention and fc layers:} 
We also seek to investigate the effectiveness of the LoRA module in conjunction with the feed-forward network, rather than the attention module only. Surprisingly, we observe a degradation in performance with the integration of LoRA into a model of this size. Specifically, we aim to incorporate LoRA with the feed-forward network ($Lf1$ \& $Lf12$) and the results of these experiments are detailed in Table \ref{tab:reg_ffn}.

\begin{table*}[t]
\fontsize{7.5}{10}\selectfont
\centering
\renewcommand{\arraystretch}{1.1}
\tabcolsep=1.5pt\relax
\begin{tabular}{l|ccc|ccc}
\hline
\textbf{\textbf{Metric}}         & \multicolumn{3}{c|}{\textbf{ACC} ($\uparrow$)}                                        & \multicolumn{3}{c}{\textbf{RMSE} ($\downarrow$)}                                       \\ \hline
\textbf{Model}                   & \multicolumn{1}{c|}{geop@500} & \multicolumn{1}{c|}{2m\_temp} & temp@850 & \multicolumn{1}{c|}{geop@500} & \multicolumn{1}{c|}{2m\_temp} & temp@850 \\ \hline
\textbf{Regional$_{Lora}$}       & \multicolumn{1}{c|}{0.582}     & \multicolumn{1}{c|}{0.823}     & 0.614   & \multicolumn{1}{c|}{438.213}    & \multicolumn{1}{c|}{1.414}     & 2.476     \\ \hline
\textbf{Regional$_{Lora+Lf1}$}   & \multicolumn{1}{c|}{0.540}     & \multicolumn{1}{c|}{0.808}     & 0.578     & \multicolumn{1}{c|}{453.996}    & \multicolumn{1}{c|}{1.459}     & 2.576      \\ \hline
\textbf{Regional$_{Lora+Lf12}$} & \multicolumn{1}{c|}{0.540}     & \multicolumn{1}{c|}{0.808}     & 0.578     & \multicolumn{1}{c|}{454.228}    & \multicolumn{1}{c|}{1.458}     & 2.574      \\ \hline
\end{tabular}
\caption{\small \textbf{\textit{LoRA module in addition to attention module}}:
Forecasting on MENA region with the regional model for 72 hrs prediction range. Resolution is $5.625^\circ$. It is observed that in addition to the attention module when Lora is extended to the feed-forward network, performance degrades.}
\label{tab:reg_ffn}
\end{table*}

\begin{table}[t]
\fontsize{7.5}{10}\selectfont
\centering
\renewcommand{\arraystretch}{1.1}
\tabcolsep=1.5pt\relax
\begin{tabular}{l|c|c|c}
\hline
\textbf{Model}             & \begin{tabular}[c]{@{}c@{}}Params\\ (M)\end{tabular} & \begin{tabular}[c]{@{}c@{}}Convergence\\ (Hrs)\end{tabular} & \begin{tabular}[c]{@{}c@{}}GPU Memory\\ (GB)\end{tabular} \\ \hline
\textbf{Regional$_{fft}$}  & 108.0                                                & $\sim$ 8.6                                                         & 15.2                                                      \\ \hline
\textbf{Regional$_{Lora}$} & 16.2                                                 & $\sim$ 4.5                                                         & 9.3                                                       \\ \hline
\end{tabular}
\caption{\small \textbf{\textit{Parameters in Lora and full fine tuning}}: Resolution is $5.652^\circ$. LoRA requires fewer parameters for training, uses less GPU memory for computations, and converges faster (in terms of training time) compared to the full fine-tuning (fft) approach.}
\label{tab:params}
\end{table}

\noindent\textbf{Memory and time comparison: } Compared to full fine-tuning, LoRA in a parameter-efficient fine-tuning paradigm, significantly reduces the number of trainable parameters and accelerates convergence. Furthermore, during training, the memory consumption on the GPU is noticeably lower in LoRA, demonstrating 38.82\% memory reduction and 85.0\% parameter reduction relative to the full fine-tuning model. For more details see Table~\ref{tab:params}.

\subsection{Qualitative}\label{figsece4}
\begin{figure*}[t]
    \centering
    \includegraphics[width=1.0\linewidth]{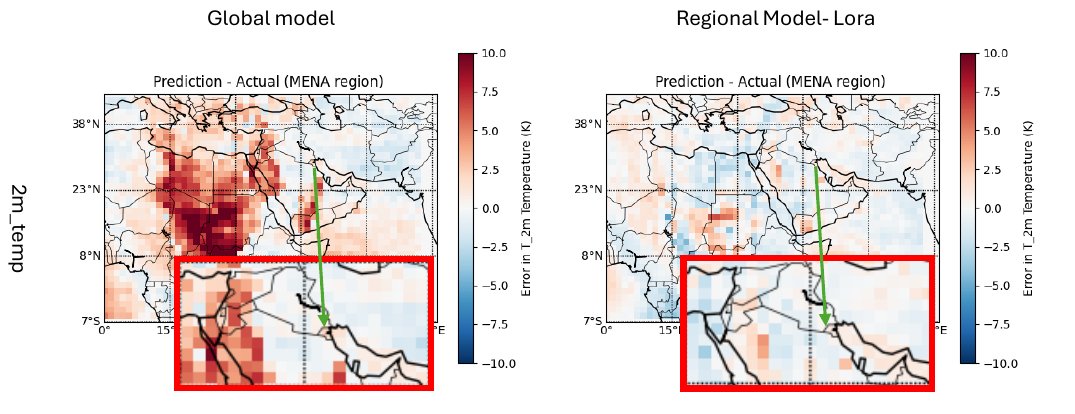}
    \caption{\small {Qualitative: Error/Bias in Predictions and Actual measurements for temperature\_2m (K), with reference to \cite{independentKuwaitSwelters}. Dated, 22$^{nd}$ June 2017}. 
    }
    \label{fig:climaxvis_kuw}
\end{figure*}

\noindent\textbf{Extreme Weather Event: }
In Fig. \ref{fig:climaxvis_kuw}, it is shown that our Lora-tuned model performs better and gives minimum error in predictions with respect to ground truth. This also signifies the stability of the model for extreme weather conditions that happened in June 2017 in Kuwait \cite{independentKuwaitSwelters}.

\begin{figure*}[t]
    \centering
    \includegraphics[width=1.0\linewidth]{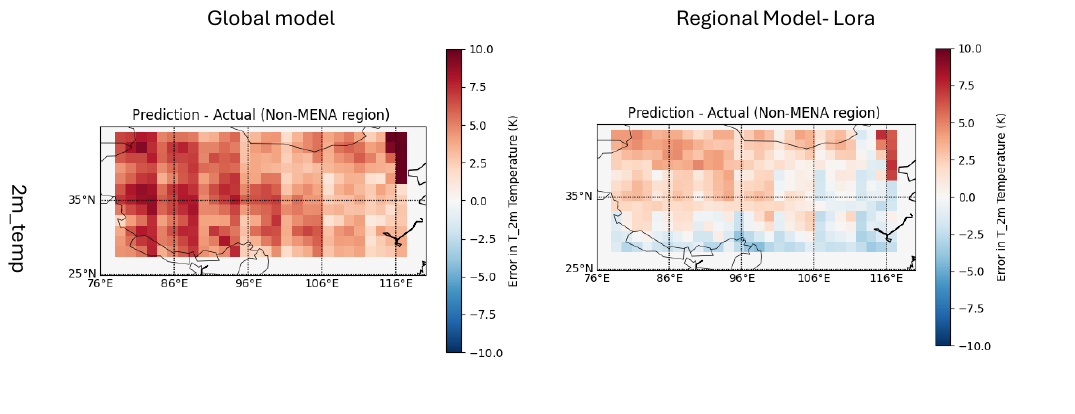}
    \caption{\small {Qualitative: Error/Bias in Predictions and Actual measurements for temperature\_2m (K), on Non-MENA region. Dated, 20$^{th}$ May 2017}. 
    }
    \label{fig:climaxvis_non}
\end{figure*}

\noindent\textbf{Non-MENA region: } To evaluate the generalizability of the regional model, we assessed its performance on the Non-MENA region (primarily China and its surroundings), where it had not encountered data over the historical range of years. Our regional model demonstrated superior performance as shown in Fig. \ref{fig:climaxvis_non}. This improved generalizability may be due to the model's focus on a specific region, allowing it to learn more consistent and relevant features and develop a coherent understanding of local data attributes. Consequently, the regional model can perform better on unseen data from other regions with similar patterns. In contrast, the global model might face challenges due to the diverse and complex data it encounters from multiple regions.

\end{document}